\pdfoutput=1

\documentclass[11pt]{article}

\usepackage{ACL2023}

\usepackage{times}
\usepackage{latexsym}

\usepackage[T1]{fontenc}

\usepackage[utf8]{inputenc}

\usepackage{microtype}

\usepackage{inconsolata}

\usepackage{float}
\usepackage{chngcntr}
\usepackage[super]{nth}
\usepackage{nicefrac}
\usepackage{microtype}

\usepackage{amsmath,amssymb,amsfonts}
\usepackage{graphicx}
\usepackage{textcomp}
\usepackage{xcolor}
\usepackage{lipsum}

\usepackage{microtype}
\usepackage{booktabs} 
\usepackage{vcell}
\usepackage{amsmath}
\usepackage{amssymb}

\usepackage{graphicx}
\usepackage{graphics}

\usepackage{multirow}
\usepackage{booktabs}
\usepackage{colortbl}
\usepackage{graphicx}
\usepackage{subfigure}
\usepackage{lipsum}
\usepackage{nicefrac}

\usepackage{float}
\usepackage{chngcntr}
\usepackage[super]{nth}
\usepackage{nicefrac}
\usepackage{microtype}

\newcommand{\comment}[1]{}

\usepackage{amsmath}

\DeclareMathOperator*{\argmax}{argmax}

\newcommand{\ie}{\textit{i.e.}}

\usepackage{hyperref}

\usepackage[ruled,vlined,linesnumbered]{algorithm2e}

\usepackage{hyperref}

\title{Driving Context into Text-to-Text Privatization}

\author{Stefan Arnold \and Dilara Yesilba \and Sven Weinzierl \\ Friedrich-Alexander-Universität Erlangen-Nürnberg \\ Lange Gasse 20, 90403 Nürnberg, Germany \\ \texttt{(stefan.st.arnold, dilara.yesilbas, sven.weinzierl)@fau.de}}

\begin{document}

\maketitle

\begin{abstract}
    \textit{Metric Differential Privacy} enables text-to-text privatization by adding calibrated noise to the vector of a word derived from an embedding space and projecting this noisy vector back to a discrete vocabulary using a nearest neighbor search. Since words are substituted without context, this mechanism is expected to fall short at finding substitutes for words with ambiguous meanings, such as \textit{'bank'}. To account for these ambiguous words, we leverage a sense embedding and incorporate a sense disambiguation step prior to noise injection. We encompass our modification to the privatization mechanism with an estimation of privacy and utility. For word sense disambiguation on the \textit{Words in Context} dataset, we demonstrate a substantial increase in classification accuracy by $6.05\%$.
\end{abstract}

\section{Introduction}
\label{sec:1}

A tension exists between the need to leverage textual data to develop language models and privacy concerns regarding the information conveyed by that data. This is of particular importance because personal information can be recovered from language models \citep{song2019auditing, carlini2020extracting, pan2020privacy}. 

Metric Differential Privacy provides a protection against the disclosure of private information. It has recently been tailored to textual analysis in the form of a text-to-text privatization mechanism \citep{feyisetan2020privacy}. Building on continuous-valued word embeddings, it relies on the assumption that words close in embedding space serve similar semantic and syntactic roles. This property of embeddings is exploited to replace all words in a text with substitute words given a probability that can be controlled by a noise parameter. A nearest neighbor search is employed to return a substitute word from all words in the embedding space.

A notable deficiency of word embeddings is that they assign a single representation to each word. Depending on its context, an ambiguous word can refer to multiple, potentially unrelated, meanings. Word embeddings are unable to reflect this dynamic nature of words, leading to potentially inappropriate substitutions when used for text-to-text privatization. Clues signaled by inappropriate substitute words may direct a classifier into the opposite direction during downstream tasks. Contextualised word embeddings are an attempt at addressing this limitation by computing dynamic representations for words which can adapt based on context. However, this dynamic behavior makes it virtually impossible to return a substitute word as the nearest neighbor search requires all vectors to be pre-computed and located in the same embedding space. 

Sense embeddings represent a middle course between lexical embeddings and contextualized embeddings. By decoupling the static representations of words into multiple representations that capture the meaning of words (covering one representation for each meaning of a word), sense representations enable context-aware text-to-text privatization. 

We make the following contributions:

\begin{enumerate}

    \item[$\bullet$] We replace the word embedding in \citet{feyisetan2020privacy} with a sense embedding constructed according to \citet{pelevina2017making}. To utilize the decoupled senses of words, we further incorporate a word-sense disambiguation prior to the privatization step that discriminates a sense given a sense inventory and a context window. 
    
    \item[$\bullet$] We investigate the privacy and utility of substitutions compared to the baseline privatization mechanism without context awareness. Congested by additional representations for each sense of a word, we find that the plausible deniability (acting as our proxy for privacy) is shaped almost identical but allows for smaller noise injection. To demonstrate the utility, we obtain substitutions of identical words paired in either the same or different contexts. At equivalent levels of privacy, the similarity of substitutions for which their original words belong to the same context show a significantly higher similarity than those of substitutions for which their original words belong to different contexts. Using a set of benchmark tasks from \texttt{GLUE} \citep{wang2019glue}, we demonstrate that this difference is an important signal for downstream classification.

\end{enumerate}

\section{Preliminaries} \label{sec:2}

\subsection{Differential Privacy}

\textit{Metric Differential Privacy} \citep{chatzikokolakis2013broadening} is a generalization of differential privacy that originated in the context of location-based privacy, where locations close to a user are assigned with a high probability, while distant locations are given negligible probability. Using word embeddings as a corollary to geo-location coordinates, metric differential privacy has been adopted from location analysis to textual analysis by \citet{feyisetan2020privacy}. This avoids the curse of dimensionality arising from randomized response \citep{warner1965randomized}. 

We follow the formulation of \citet{xu2021utilitarian} for metric differential privacy in the context of textual analysis. Equipped with a discrete vocabulary set $\mathcal{W}$, an embedding function $\phi : \mathcal{W} \rightarrow \mathbb{R}$, where $\mathbb{R}$ represents a high-dimensional embedding space, and a distance function $d: \mathbb{R} \times \mathbb{R} \rightarrow [0,\infty)$ satisfying the axioms of a metric (\ie, identity of indiscernibles, symmetry, and triangle inequality), metric differential privacy is defined in terms of the distinguishability level between pairs of words. Formally, a randomized mechanism $\mathcal{M}:\mathcal{W} \rightarrow \mathcal{W}$ satisfies metric differential privacy with respect to the distance metric $d(\cdot)$ if for any $w,w^{'},\hat{w} \in \mathcal{W}$ the distributions of $\mathcal{M}(w)$ and $\mathcal{M}(w^{'})$ are bounded by Equation \ref{equation:dp2} for any privacy budget $\varepsilon > 0$:

\begin{equation}
\label{equation:dp2}
 \frac{\mathbb{P} [\mathcal{M}(w) = \hat{w}]}{\mathbb{P} [\mathcal{M}(w^{'}) = \hat{w}]} \leq e^{\varepsilon d\{\phi(w),\phi(w^{'})\}}.
\end{equation}

This probabilistic guarantee ensures that the log-likelihood ratio of observing any word $\hat{w}$ given two words $w$ and $w’$ is bounded by $\varepsilon d\{\phi(w),\phi(w’)\}$, providing plausible deniability \citep{bindschaedler2017plausible} with respect to all $w \in \mathcal{W}$. We refer to \citet{feyisetan2020privacy} for a complete proof of privacy. For the mechanism $\mathcal{M}$ to provide plausible deniability, additive noise is in practice sampled from a multivariate distribution such as the \textit{multivariate Laplace distribution} \citep{feyisetan2020privacy} or \textit{truncated Gumbel distribution} \citep{xu2020adversarial}.

We recall that differential privacy requires adjacent datasets that differ in at most one record. Since the distance $d(\cdot)$ captures the notion of closeness between datasets, metric differential privacy instantiates differential privacy when Hamming distance is used, \ie, if $\forall x,x': d\{\phi(w),\phi(w^{'})\} = 1$. Depending on the distance function $d(\cdot)$, metric differential privacy is therefore generally less restrictive than differential privacy. Intuitively, words that are distant in metric space are easier to distinguish compared words that are in close proximity. Scaling the indistinguishability by a distance $d(\cdot)$ avoids the curse of dimensionality that arises from a large vocabulary $\mathcal{W}$ and allows the mechanism $\mathcal{M}$ to produce similar substitutions $\hat{w}$ for similar $w$ and $w^{'}$. However, this scaling complicates the interpretation of the privacy budget $\varepsilon$, as it changes depending on the metric employed.

\paragraph{Related Work.} The multivariate mechanism for text-to-text privatization by \citet{feyisetan2020privacy} has been extended in orthogonal directions to further improve the utility \citep{feyisetan2019leveraging, carvalho2021tem} and privacy \citep{xu2020differentially}. 

Drawing inspiration from \citet{feyisetan2019leveraging}, we complement on the line of inquiry dedicated to the enhancement of the utility. By leveraging the curvature of the space at different locations in the Hyperbolic space of Poincaré embeddings \citep{nickel2017poincare}, their mechanism preserves the hierarchical structure of words during substitution. We persist in the Euclidean space and instead replace the word embedding with a sense embedding to account for the ambiguity of words during substitution. Our results demonstrate that this modification leads to improved performance on downstream tasks while being compatible with prevalent embedding mechanisms.

\subsection{Word Embeddings}

Since metric differential privacy for text-to-text privatization operates on word embeddings, the merits of privatization are limited by the capabilities of these word embeddings. Starting from sparse vectors suffering from curse of dimensionality, which makes computation and storage infeasible, most research on word embeddings is dedicated to learning dense vectors from corpus-level co-occurrence statistics \citep{mikolov2013efficient}. To learn these dense vectors, two mirrored approaches have been proposed: continuous bag-of- words and skip-gram. Continuous bag-of- words is trained to predict a word from a fixed window size of context words, whereas skip-gram specifies the probability of observing the context words conditioned on a word within a window. This results in a real-valued vector representation of words that capture interpretable analogical relations between words.

A limitation of these embedding mechanisms is that they conflate all meanings of a word into a single representation, and the most frequent meaning of a word dominates this representation. By conflating all meanings, word embeddings are unable to discriminate ambiguous words. This inability to distinct between ambiguous words is inherited to word substitutions obtained from privatization.

\subsection{Sense Embeddings}

To address the meaning conflation deficiency of word embeddings, one can represent meanings of words in the form of sense embeddings. Learning sense embeddings has been an active area of research until the emergence of contextual embeddings. We briefly recall some methods to sense representation. Exploiting an unlabeled corpus of text, methods to resolve the meaning conflation deficiency can be divided into three main branches: (1) a staged induction of word senses followed by learning of sense representations, (2) a joint induction of word senses together with learning of sense representations, and (3) retrofitting an existing word embedding by de-conflating word representations into sense representations. 

The sense distinctions required to discriminate the meaning of a word are extracted from text corpora by clustering words according to their contexts given a window size. This paradigm is related to word-sense induction. It comes with algorithmic complexity and interpretability problems. Instead of a word-sense induction by clustering, an alternative approach is to derive word senses from pre-defined sense inventories. This paradigm is related to word-sense disambiguation in which ambiguous words must be assigned a sense from the sense inventory. Exploiting knowledge from pre-defined sense inventories for the initialization of senses allows learning representations that are linked to interpretable sense definitions. Two shortcomings are apparent to learning sense representations using word-sense disambiguation. It is assumed that the sense distinctions intended by the text matches those defined in the sense inventory. Unable to handle words that are not defined in the sense inventory, relying on pre-defined senses hinges on the coverage of the sense inventory.

\paragraph{Staged training of sense embeddings.} The training of sense embeddings initially employed a staged approach \citep{reisinger2010multi, huang2012improving, vu2016k}. \citet{reisinger2010multi} constructed sense vectors by clustering sparse vectors corresponding to occurrences of words into a predetermined number of clusters. Clustering is performed by a parametric method that permits controlling the semantic breadth using a per-cluster concentration. Assuming a fixed fixing number of senses for all words, the centroids of the clusters are used as sense vectors and word occurrences are relabeled according to the cluster they belong to. This idea has been extended to dense vectors \citep{huang2012improving}.

Instead of inducing senses by clusters, a straightforward method is to disambiguate text corpora as defined by a sense inventory and apply an embedding method on the resulting sense-annotated text \citep{iacobacci2015sensembed, flekova2016supersense, ruas2019multi}. \citet{iacobacci2015sensembed}, for instance, use an off-the-shelf disambiguation process to obtain a sense-annotated corpus and directly learn sense representations.

\paragraph{Joint training of sense embeddings.} A staged approach to learning sense representations suffers from the limitation that clustering and learning does not take advantage from their inherent similarities. To avoid the issues brought by a two-step clustering, the idea of clustering context vectors has been adapted into the training of word embeddings \citep{ tian2014probabilistic, pina2014simple, neelakantan2014efficient,  liu2015topical, liu2015learning, bartunov2016breaking, lee2017muse, nguyen2017mixture}. Performing clustering and embedding learning jointly, the intended sense for each word is dynamically selected as the closest sense to the context and weights are updated only for that sense. Assuming a fixed number of senses per word, \citet{tian2014probabilistic} introduced an expectation maximization integrated with skip-gram that learns multiple senses weighted by their prior probability. Since words can have a highly dynamic number of senses that range from monosemous words to polysemous words with dozens of associated meanings, this assumption presents a severe limitation. \citet{pina2014simple} address the varying polysemy problem of sense representation by setting the number of senses of a word as defined by a sense inventory. Deriving the number of senses for each word from a sense inventory, it does not need to create or maintain clusters to discriminate between senses. A better solution would involve dynamic induction of senses from the text corpus. \citet{neelakantan2014efficient} applies a non-parametric clustering procedure for estimating the granularity of senses for each word. Similar to \citet{tian2014probabilistic}, it represents the context of a word as the centroid of the vectors of its words but allocates a new sense vector each time the similarity of a context to existing senses is below a certain threshold. By using latent topic modeling to assign topics to each word in a corpus \citep{liu2015topical,liu2015learning} and a mixture of weights that reflect different association degrees of each word to multiple senses in the context \citep{nguyen2017mixture}, words can be discriminated into more general topics. 


\paragraph{Retrofitting of word embeddings.}

Instead of training a word and sense embedding jointly, research exists on refining a word embedding to match semantic constraints \citep{faruqui2014retrofitting, jauhar2015ontologically, johansson2015embedding, rothe2015autoextend, collier2016conflated}. Given a word embedding, \citet{faruqui2014retrofitting} propose \textit{retrofitting} as a post-processing step in which words that are connected by a relationship derived from a semantic network are moved closer together in the embedding space. \citet{jauhar2015ontologically} tailored retrofitting towards learning representations for the senses listed in a sense inventory. Using a random walk, \citet{collier2016conflated} extracted a set of sense biasing words from an external sense inventory. To de-conflate a word, they add a set of sense embeddings to the same space and push words in the space to the region occupied by its corresponding sense biasing words.

Most retrofitting approaches rely on signals from sense inventories. To transform word embeddings to sense embeddings without external resources, \citet{pelevina2017making} construct a graph by connecting each word to a set of related words. Using ego-network clustering of words, senses are induced as a weighted average of words in each cluster. 

\subsection{Contextual Embeddings}

Although much research has been directed to sense embeddings, the field shifted towards learning contextual embeddings \citep{peters2018deep, devlin2019bert}. Rather than pre-computing a static representation for each word, contextualized embeddings dynamically change the representation of a word depending on the context. Harnessing sense signals during the training objective of contextual embeddings has been shown to promote the disambiguation of word meanings \citep{peters2019knowledge,huang2019glossbert,levine2020sensebert, scarlini2020sensembert}. However, the dynamic representations produced by contextual embeddings disqualifies contextual embeddings for privatization as the nearest neighbor search requires that the representations are aligned in a shared embedding space.


\section{Methodology} \label{sec:3}

Aiming at context-aware privatization of ambiguous words in texts, we adopt the privatization mechanism of \citet{feyisetan2020privacy} and replace the word embedding with a sense embedding. The sense embedding is constructed by building and clustering a graph of nearest neighbors based on vector similarities \citep{pelevina2017making}.

Using a context window of size $3$ and minimum word frequency of $5$, we construct a $300$-dimensional word embedding on a dump of \texttt{Wikipedia}. We align our vocabulary with words contained in \texttt{GloVe}. Our word embedding contains $95,670$ words with words vectors. For each word in the word embedding, we retrieve its $200$ nearest neighbors according to the cosine similarity of their word vectors. Once calculated the similarities, we build a graph of word similarities. Assuming that words referring to the same sense tend to be tightly connected, while having fewer connections to words referring to different senses, word senses can be represented by a cluster of words. 

A sense inventory is induced from ego-network clustering. The clustering yielded $248,218$ word senses. Each word sense is indexed by a sense identifier. Performing graph clustering of ego-networks is non-parametric. It makes no assumptions about the number of word senses. However, the number and definition of the resulting word senses are not linked to a lexical inventory. Since a word sense is assumed as a composition of words in a cluster, sense vectors are calculated as a weighted pooling of word vectors representing cluster items. 

\begin{figure}[thp]
    \includegraphics[width=0.45\textwidth]{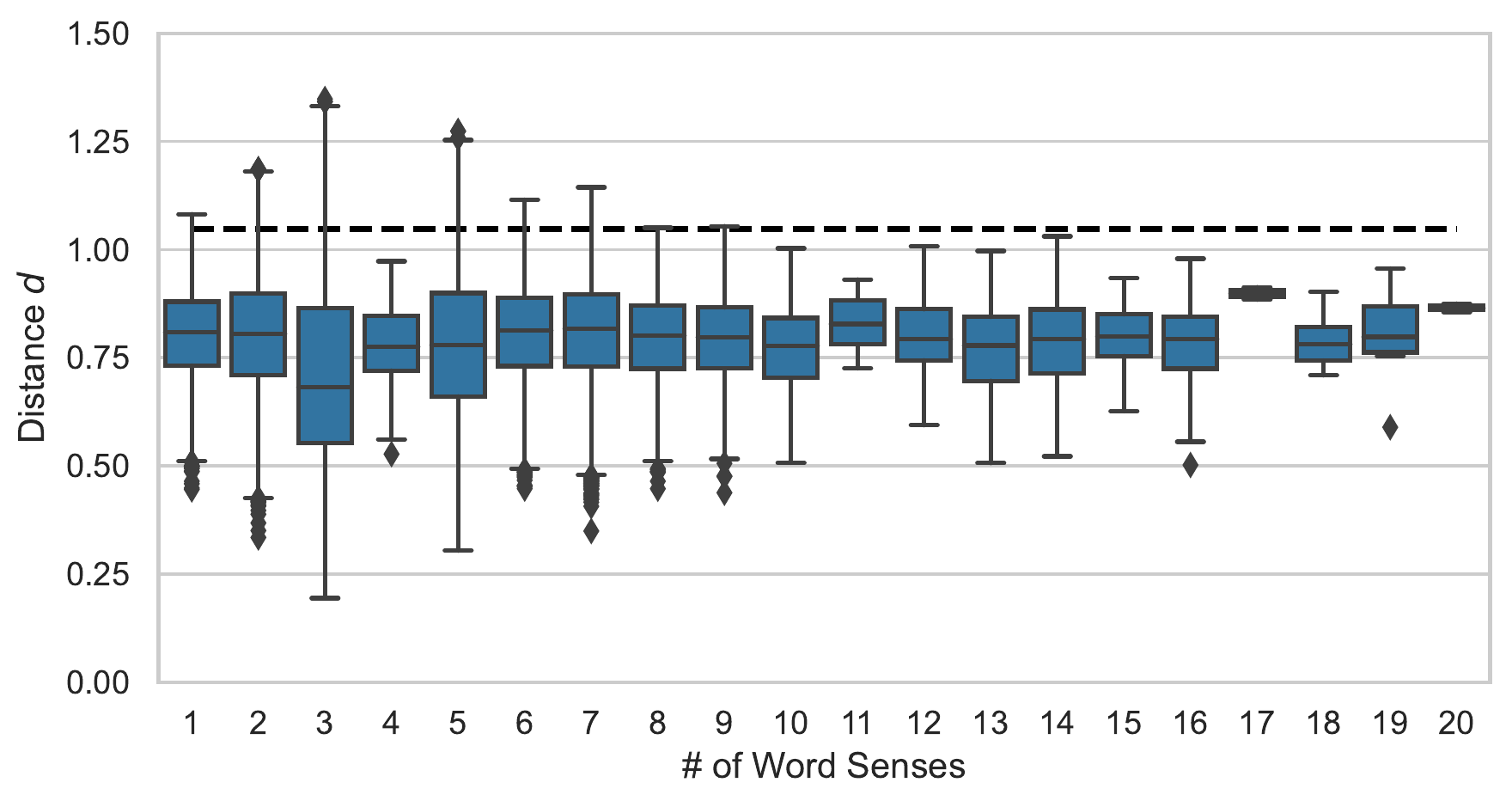}
    \caption{Pairwise Euclidean distances within word senses as a function of the number of distinct senses. The dashed line corresponds to the averaged pairwise distance of word forms in the embedding space.}
    \label{fig:distances}
\end{figure}

In Figure \ref{fig:distances}, we depict the averaged pairwise distances of words as a function of the number of senses. On average, the distance within word senses is considerably lower than the average distance between words in the embedding space (depicted by a dotted line at $1.0550$). Since the privatization step is applied directly to the structure of the embedding space, the distance between senses originating from the same word must be taken into account when assessing utility and privacy. 

To utilize the sense representations, we incorporate a disambiguation step prior to the privatization. Given a word and its context words, we map the word to a set of its sense vectors according to the sense inventory.  The disambiguation strategy is based on similarity between sense and context words: $\argmax \nicefrac{\overline{\mathbf{c}} \cdot \mathbf{s}_i} {\|\overline{\mathbf{c}}\| \cdot \|\mathbf{s}_i\|}$, where $\overline{\mathbf{c}}$ is the mean of the word vectors from the context words. In line with the context size during sense induction, context words for the sense disambiguation are selected within a window of $5$. This step is repeated for each word prior to the privatization step. 

The privatization step follows a multi-step protocol: We retrieve the sense vector for each disambiguated word. This sense vector is perturbed with noise sampled from a multivariate distribution and its noisy representation is then projected back to the discrete vocabulary space of the sense embedding. As noisy representations are unlikely to exactly represent words in the embedding space, a nearest neighbor approximation is returned. To obtain a private text of word forms, we truncate the sense identifier from the word senses. The result is a privatized text that can be post-processed by word embeddings agnostic to the sense embedding.

\begin{figure}[t]
    \subfigure[Lexical substitutions for \textit{'bank'}]
    {
        \fbox{\includegraphics[width=0.45\textwidth]{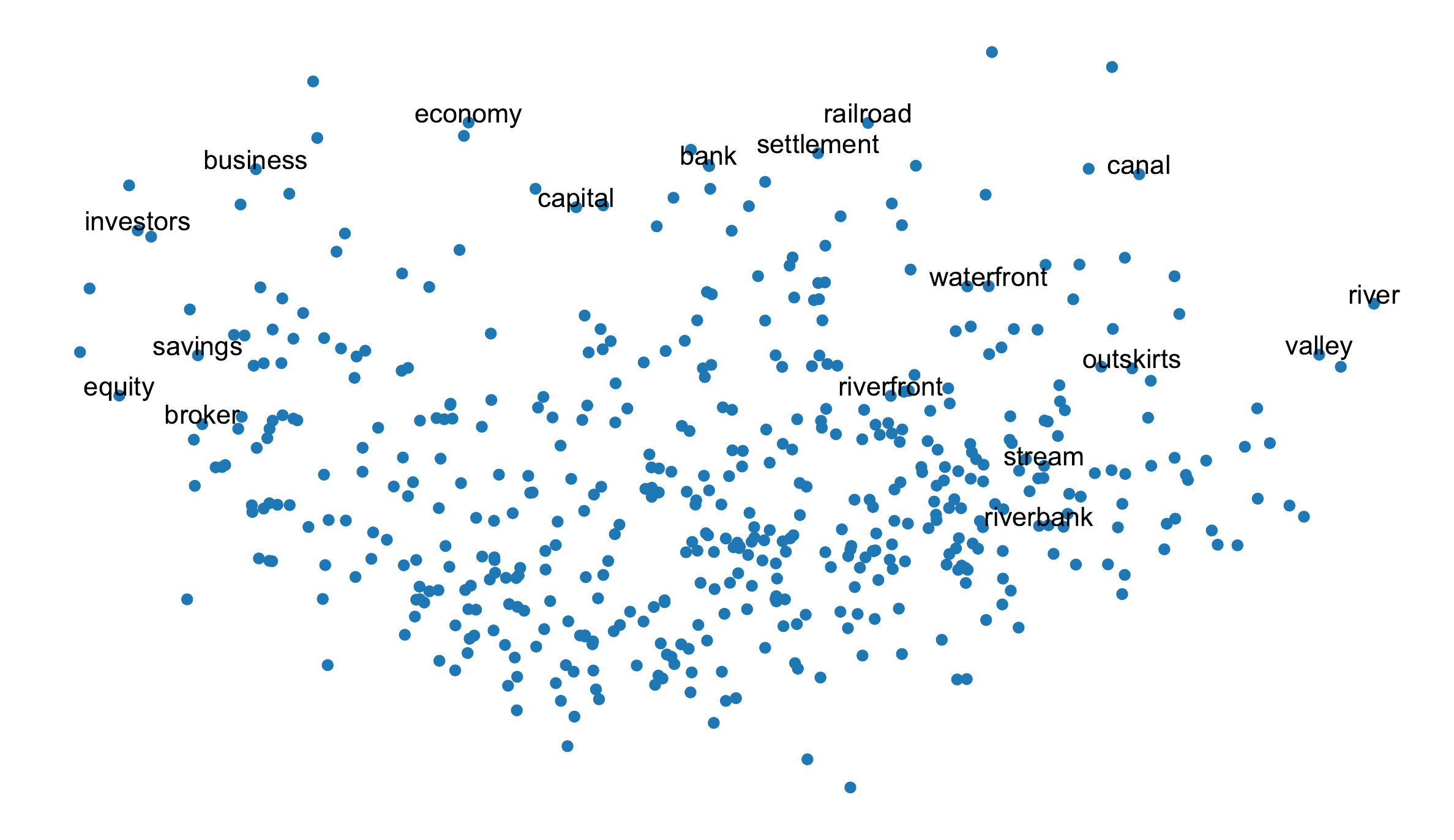}}
        \label{fig:lex_example}
    }
    \hfill
    \subfigure[Contextual substitutions for \textit{'bank'}]
    {
        \fbox{\includegraphics[width=0.45\textwidth]{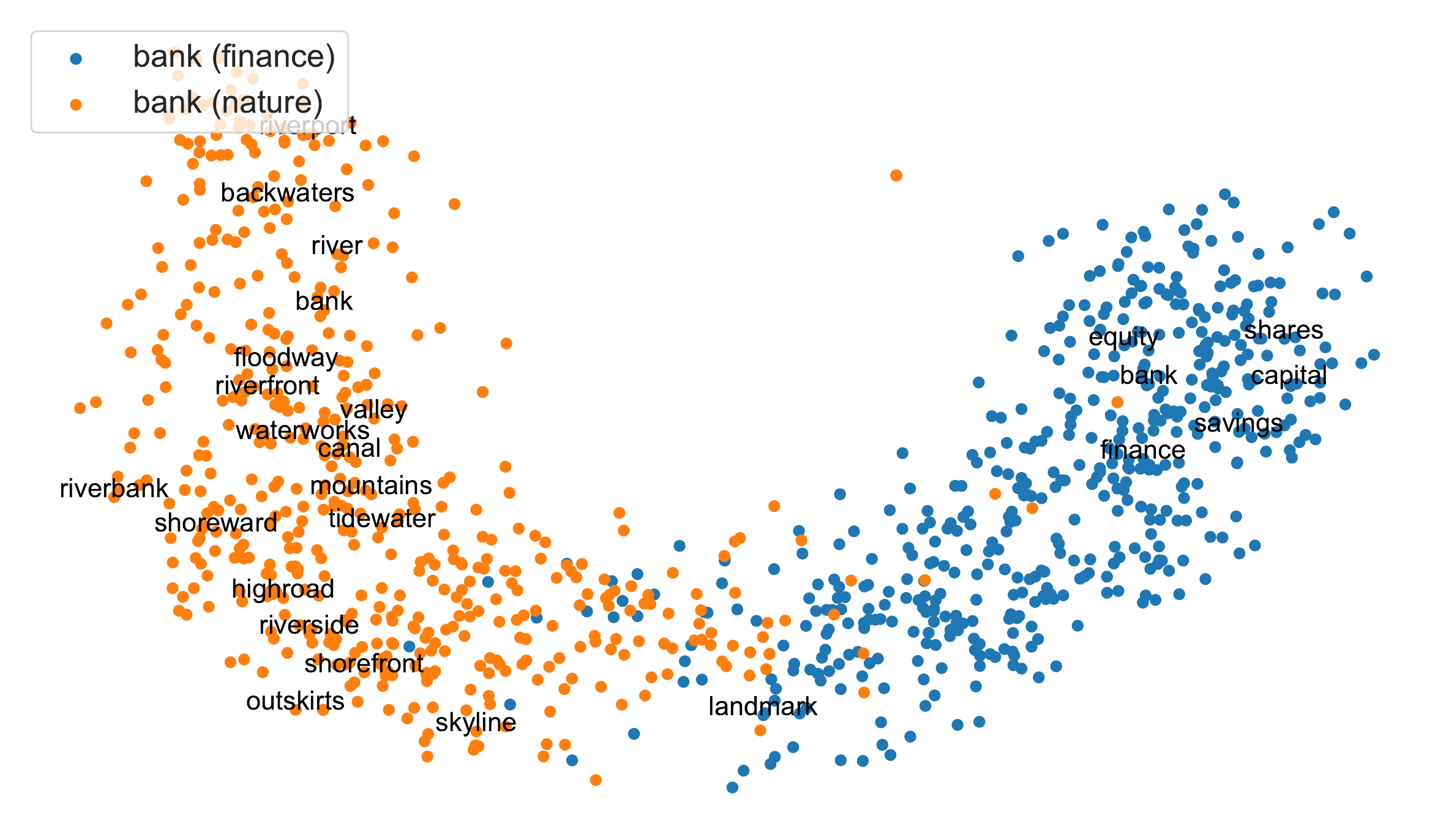}}
        \label{fig:ctx_example}
    }
    \caption{Example substitutions associated with a geographical and financial context. A seamless transition in Figure \ref{fig:lex_example} compared to distinct regions in Figure \ref{fig:ctx_example}.}
    \label{fig:example}
\end{figure}

To demonstrate the effectiveness of leveraging sense embedding in combination with a disambiguation step prior to the privatization, we privatized the ambiguous word \textit{'bank'} for a total of $500$ queries and recorded its substitutions. In half of the queries, the ambiguous word is contained in a text belonging to a geographical context, and in the other half, the ambiguous word is contained in a text belonging to a financial context. The texts are \textit{'to walk by a river \textbf{bank} at sunset'} and \textit{to deposit money at a \textbf{bank} to earn interest'}. We reduced the dimensionality of the substitute vectors into a two-dimensional space for visualization in Figure \ref{fig:example}. We highlight words of the obtained substitutions. We observe that the substitution words returned by lexical privatization stem from both geographical and financial contexts. While substitutions blend between senses during lexical privatization, we discover distinct boundaries between substitute words belonging to contrasting contexts if the words are disambiguated before privatization. 

\section{Experiments}

\subsection{Privacy Analysis}

The privacy guarantees in metric differential privacy depend on the deployed metric and the geometric properties of the embedding space. Since retrofitting changes the geometric properties by populating the geometric space of the embedding with word senses that refer to the same word form, we need to recalibrate the plausible deniability \citep{bindschaedler2017plausible}. We record the following statistics as proxies for the plausible deniability. We note that these proxy statistics have been used in previous studies to characterize the plausible deniability of multivariate mechanisms \cite{feyisetan2019leveraging, feyisetan2020privacy, xu2020differentially, xu2021utilitarian}.

\begin{enumerate}

    \item[$\bullet$] $N_w = \mathbb{P} \{ M(w) = w \}$ measures the probability that a word is not substituted by the mechanism. This is approximated by counting the number of occurrences in which a word $w$ is substituted by the same word after running the mechanism for $100$ times.
    
    \item[$\bullet$] $S_w = |\mathbb{P} \{ M(w) = w^{‘} \}|$ measures the effective support in terms of the number of distinct substitutions produced for a word from the mechanism. This is approximated by the cardinality of the set of words $w^{‘}$ after running the mechanism for $100$ times.
    
\end{enumerate}

Since the noise in the multivariate Laplace mechanism is scaled by $\nicefrac{1}{\epsilon}$, we can make a connection between the proxy statistics and the privacy budget $\epsilon$. A smaller $\epsilon$ corresponds to more stringent privacy guarantees by adding more noise to the word embedding. More noise leads to fewer unperturbed words (lower $N_w$) and more diverse outputs for each word (higher $S_w$). By contrast, a higher $\epsilon$ leads to less substitutions (higher $N_w$) and a narrow set of distinct words (lower $S_w$). From a distributional perspective, it follows that $N_w$ ($S_w$) should be positively (negatively) skewed to afford reasonable privacy guarantees.

\begin{figure}[t]
    \subfigure[Lexical $N_w$]
    {
         \includegraphics[width=0.45\textwidth]{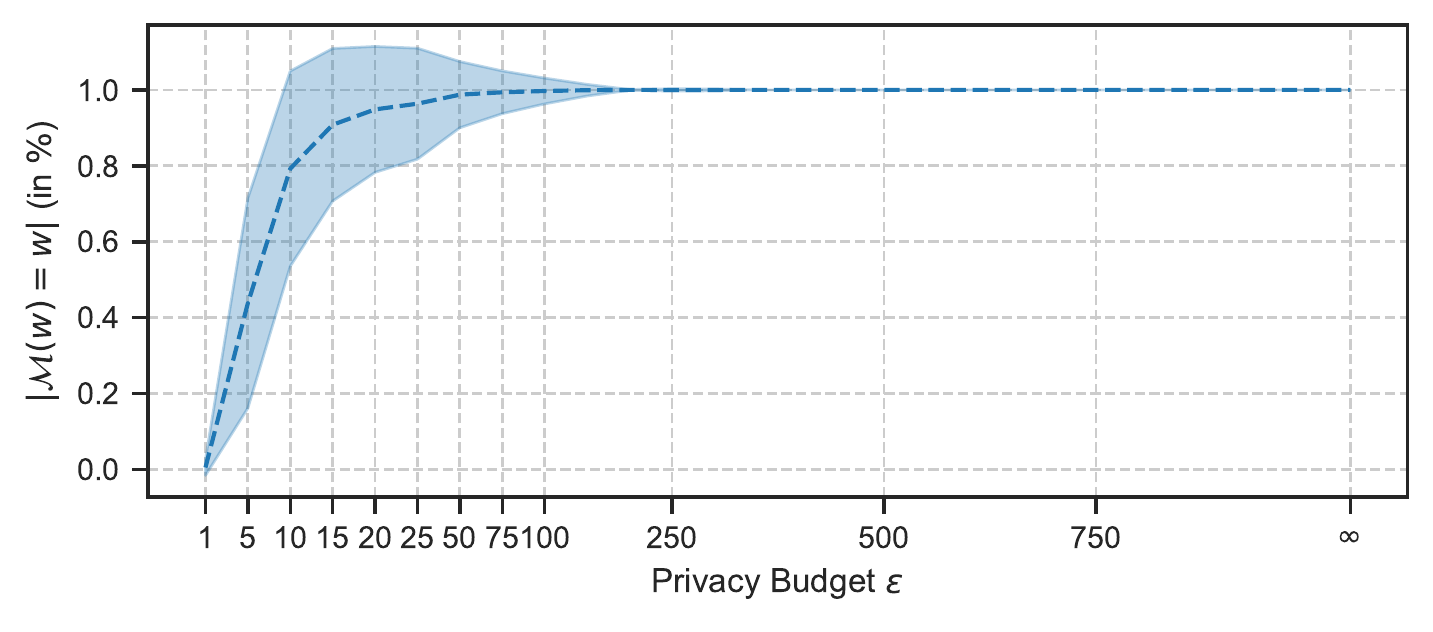}
         \label{fig:nw_lex}
    }
    \hfill
    \subfigure[Contextual $N_w$]
    {
        \includegraphics[width=0.45\textwidth]{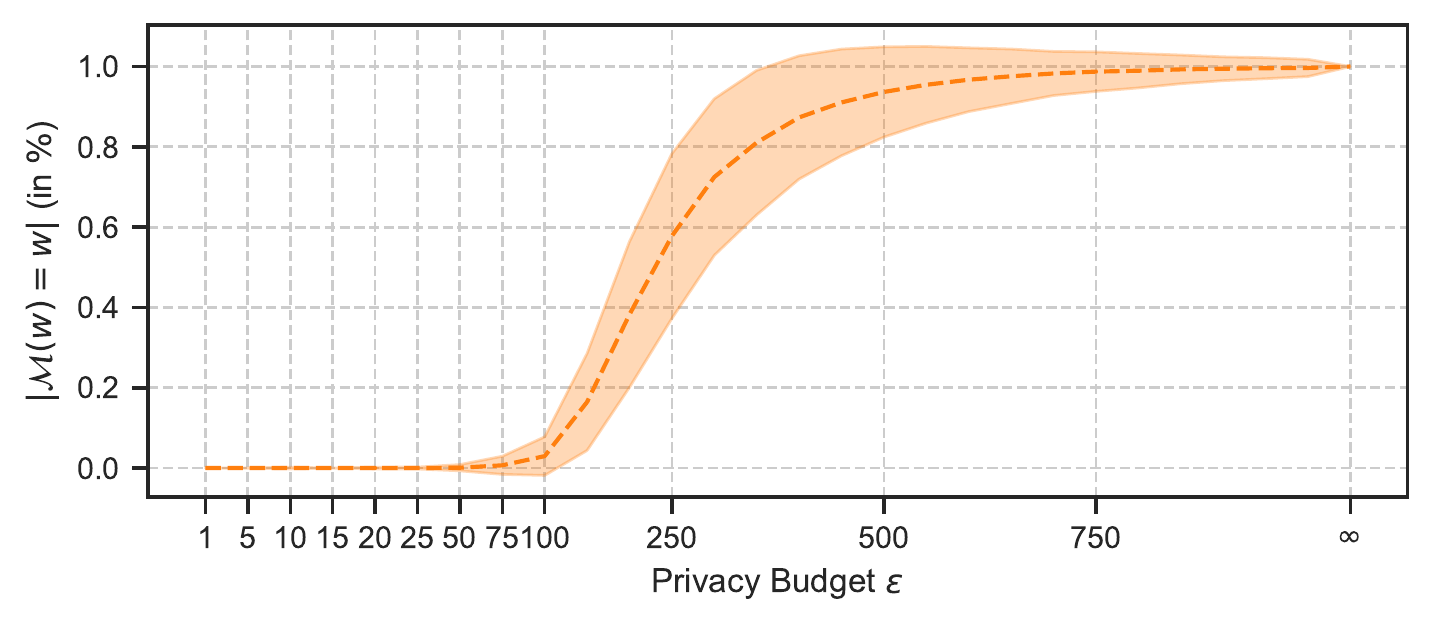}
        \label{fig:nw_ctx}
    }
    \caption{$N_w$ refers to the number of substitute words that are \textit{identical} to a queried sensitive word. The shift in the curve suggests that higher privacy budgets are legitimate before there is a risk that words will not be replaced by substitutions.}
    \label{fig:nw}
\end{figure}

\begin{figure}[t]
    \subfigure[Lexical $S_w$]
    {
         \includegraphics[width=0.45\textwidth]{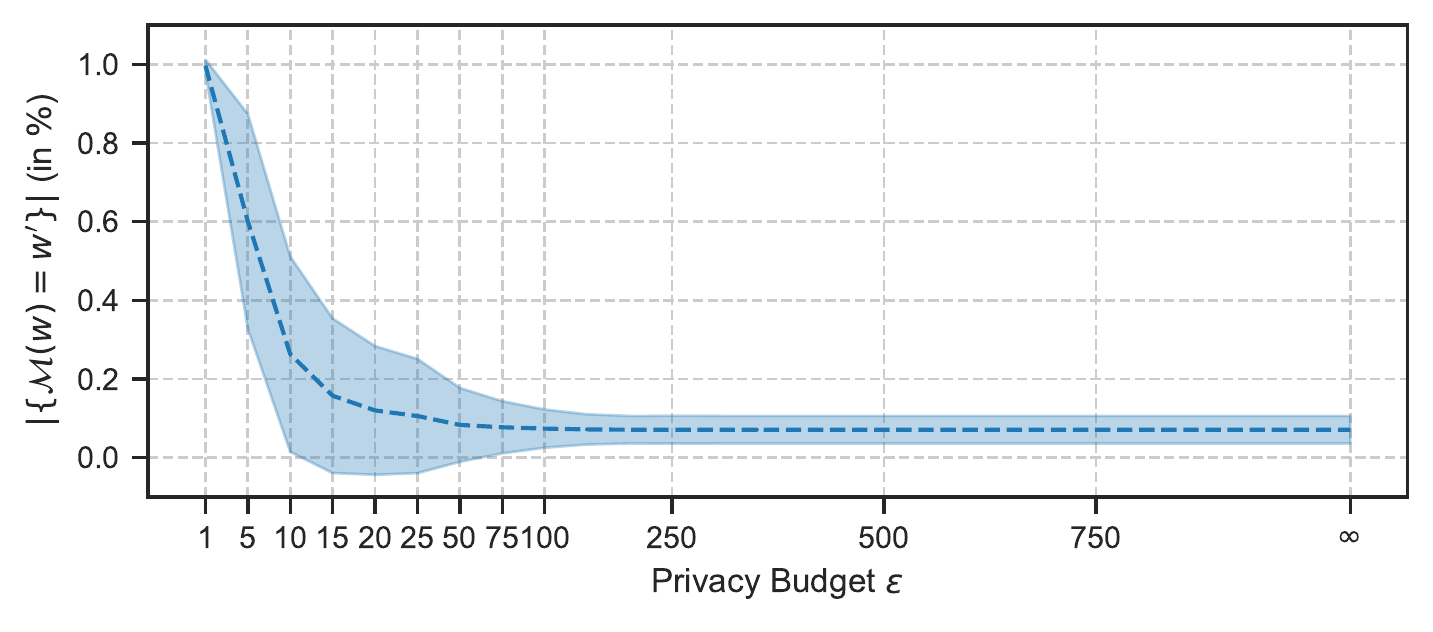}
         \label{fig:sw_lex}
    }
    \hfill
    \subfigure[Contextual $S_w$]
    {
        \includegraphics[width=0.45\textwidth]{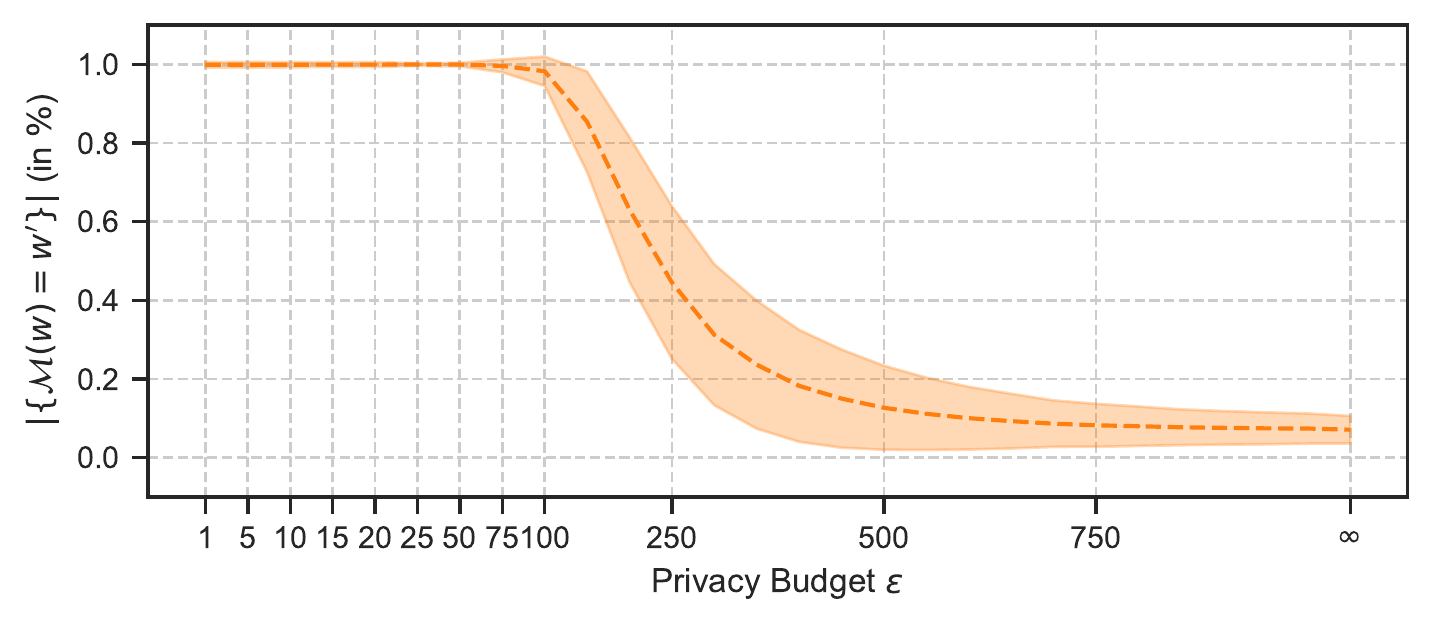}
        \label{fig:sw_ctx}
    }
    \caption{$S_w$ refers to the number of substitute words that are \textit{unique} from a queried sensitive word. The shift in the curve suggests that higher privacy budgets are legitimate before the effective support of substitution candidates violates plausible deniability.}
    \label{fig:sw}
\end{figure}

In Figures \ref{fig:nw} and \ref{fig:sw}, we present the averaged values of $N_w$ and $S_w$ over $100$ independent queries from the corpus of \texttt{WikiText} \citep{merity2016pointer} for a discrete set of privacy budgets $\varepsilon = \{1,5,10,15,25,50,100,250,500,\infty \}$. While lower values of $\varepsilon $ are desirable in terms of privacy, plausible deniability is assured unless $N_w$ ($S_w$) exceeds (falls below) $0.5$. The plots thus serve as a visual guidance for comparing (and selecting) the privacy budget $\varepsilon $. The curve of the privacy proxies as function of the privacy budget is shaped identical for word and sense embeddings, except that using a sense embedding stretches the allocatable privacy budget by an order of magnitude. We attribute this shape to the congestion of the embedding space with substitution candidates, even at low levels of noise.

For our utility experiments, we set the privacy budget for each mechanism so that $.90$ quantile of words is plausible deniable. To calculate the $.90$ quantile, we interpolated the scores for $N_w$ ($S_w$) and selected the privacy budget $\varepsilon$ so that $N_w$ ($S_w$) does not exceed (fall below) $0.5$. A plausible deniability for only a quantile of words was also assumed in a prior study by \citet{xu2020differentially}.

\subsection{Utility Analysis}

To analyze the utility of privatization with context awareness, we use the standard datasets for evaluating word similarity. The datasets include \texttt{WordSim-353} \citep{agirre2009study}, \texttt{SimLex-999} \citep{hill2015simlex}, and \texttt{SWCS} \citep{huang2012improving}. Common to all these datasets is that similarity ratings are given to pairs of words. While \texttt{WordSim-353} and \texttt{SimLex-999} provide pairs of words in isolation, \texttt{SWCS} provides a context for each word that triggers a specific meaning, making it very suitable for the evaluation of context-aware privatization. All experiments are conducted while ensuring plausible deniability for $.90$-quantile of words.

We query each pair of words $(w_i, w_j)$ for $25$ times by each privacy mechanism and record their similarity after privatization. We use the cosine distance as our similarity measure. The results capture $\nicefrac{\mathbf{\hat{w}}_i \cdot \mathbf{\hat{w}}_j}{\|\mathbf{\hat{w}}_i \| \cdot \|\mathbf{\hat{w}}_j \|}$. Once queried, we correlate the measured similarity against the similarity annotations. We present the results in Table \ref{tab:utility}. Without a context provided to discriminate a word, the privatisation using sense embeddings generalizes to privatisation using word embeddings. This can be seen by the almost identical correlation coefficients for \texttt{WordSim-353} and \texttt{SimLex-999}. The correlation of the sense embedding surpassing those for the word embedding on \texttt{SWCS} indicates that the information provided by the disambiguation step helps in finding more appropriate substitutions. 

\begin{table}
    \centering
    \begin{tabular}{llc|cc} 
    \toprule
                                  &  & $(w_i, w_j)$ & \textbf{Words} & \textbf{Senses}  \\ 
    \midrule
    \textbf{\textbf{WordSim-353}} &  & 0.5849         & 0.1353         & 0.1478         \\
    \textbf{\textbf{SimLex-999}}  &  & 0.2978         & 0.0696         & 0.0841         \\
    \textbf{\textbf{SCWS}}        &  & 0.5183         & 0.1911         & 0.2358         \\
    \bottomrule
    \end{tabular}
    \caption{Datasets for measuring the similarity between words. Similarity measured after substitution. Scores denote the correlation compared to annotations.}
    \label{tab:utility}
\end{table}

We further benchmark our mechanism in combination with a \texttt{BERT} model for downstream classification. We employ the words in context \citep{pilehvar2019wic} dataset. It is composed of $5,428$ text-pairs for training and $638$ text-pairs for validation. Framed as a binary classification task, the goal of words in context is to identify if the occurrences of a word for which two contexts are provided correspond to the same intended meaning. Each of context is designed to trigger a specific meaning. Note that the dataset is balanced, hence, a context-insensitive embedding would perform similarly to a random baseline.

Without privacy guarantees, \texttt{BERT} peaks at an accuracy score of $0.6887$. The training using the privatized data mimics the training without privatization. After privatizing the training data using word embeddings, \texttt{BERT} scores $0.6006$. Leveraging sense embeddings, we boost the accuracy to $0.6423$. This narrows the gap in accuracy by $6.05\%$. All scores are calculated as an average over three independent trials for each privatization mechanism. 

\begin{figure}[t]
    \centering
    \includegraphics[width=0.45\textwidth]{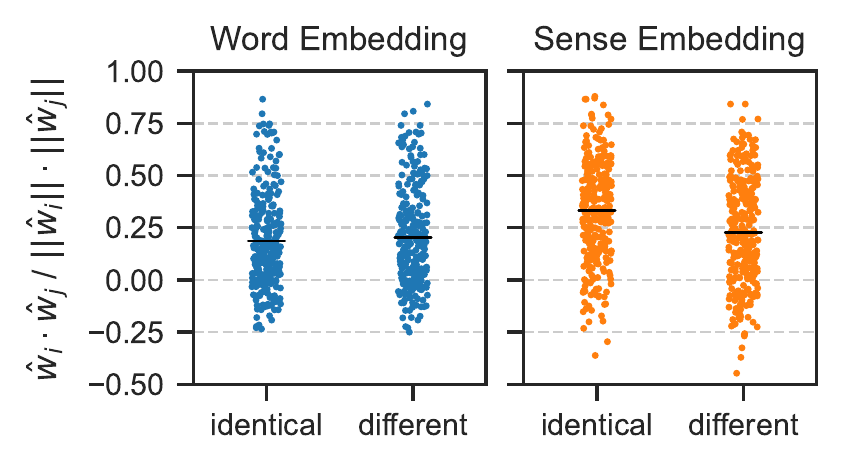}
    \caption{Cosine similarity of word pairs after substitution. The vertical line represents the average similarity.}
    \label{fig:utility}
\end{figure}

\begin{table*}
\centering
\caption{My caption}
\label{my-label}
\resizebox{\linewidth}{!}{%
\begin{tabular}{ccllllllllllll} 
\toprule & & \multicolumn{2}{c}{\textbf{Classification}} & \multicolumn{1}{c}{} & \multicolumn{3}{c}{\textbf{Textual Similarity}} & \multicolumn{1}{c}{} & \multicolumn{3}{c}{\textbf{Textual Entailment}} & \multicolumn{1}{c}{} & \multicolumn{1}{c}{\textbf{Avg}.} \\ 
\cline{3-4}\cline{6-8}\cline{10-12}\cline{14-14} &

\multirow{2}{*}{\begin{tabular}[c]{@{}c@{}}\textbf{Level of} \\ \textbf{Privacy}\end{tabular}} & \multicolumn{1}{c}{CoLA}  & \multicolumn{1}{c}{SST2}  & \multicolumn{1}{c}{} & \multicolumn{1}{c}{QQP}   & \multicolumn{1}{c}{MRPC}  & \multicolumn{1}{c}{STSB}  & \multicolumn{1}{c}{} & \multicolumn{1}{c}{MNLI}  & \multicolumn{1}{c}{QNLI}  & \multicolumn{1}{c}{RTE}   & \multicolumn{1}{c}{} & \multicolumn{1}{c}{\multirow{2}{*}{-}}  \\ & & \multicolumn{1}{c}{(MCC)} & \multicolumn{1}{c}{(ACC)} & \multicolumn{1}{c}{} & \multicolumn{1}{c}{(ACC)} & \multicolumn{1}{c}{(ACC)} & \multicolumn{1}{c}{(SCC)} & \multicolumn{1}{c}{} & \multicolumn{1}{c}{(ACC)} & \multicolumn{1}{c}{(ACC)} & \multicolumn{1}{c}{(ACC)} & \multicolumn{1}{c}{} & \multicolumn{1}{c}{} \\ 
\midrule
\textbf{BERT} & - & 0.5792 & 0.9243 &  & 0.8879 & 0.8329 & 0.8854 & & 0.8229 & 0.8912 & 0.6927 & & 0.8146 \\ 
\hline 
\multirow{2}{*}{\textbf{Words}}  & \multicolumn{1}{l}{p=0.9} & 0.0000 & 0.7614 & & 0.6883 & \textbf{0.6059} & 0.5619 & & 0.5270 & 0.6145 & 0.5342 & & 0.5367 \\

& \multicolumn{1}{l}{p=0.5}                                                                  & 0.0416                    & 0.8518                    &                      & 0.7858                    & 0.6123                    & 0.5907                    &                      & 0.7001                    & 0.7893                    & 0.5880                    &                      & 0.6200                                  \\ 
\hline

\multirow{2}{*}{\textbf{Senses}} & \multicolumn{1}{l}{$p=0.9$}                                                                  & 0.0000                    & \textbf{0.8669}           &                      & \textbf{0.7715}           & 0.5910                    & \textbf{0.6197}           &                      & \textbf{0.6750}           & \textbf{0.7446}           & \textbf{0.5834}           &                      & \textbf{0.6065}                         \\
                                 & \multicolumn{1}{l}{$p=0.5$}                                                                  & 0.0655                    & 0.8862                    &                      & 0.8215                    & 0.6322                    & 0.6442                    &                      & 0.7417                    & 0.8180                    & 0.6070                    &                      & 0.6520                                  \\

\bottomrule
\end{tabular}%
}
\caption{Results on a subset of \texttt{GLUE} \citep{wang2019glue}. We report Matthews correlation for the \texttt{CoLA} dataset, Spearman correlation for the \texttt{STSB} dataset, and the accuracy score for all remaining datasets. The level of privacy increases with the quantile of words that are provable plausible deniable. $p=.90$ denotes an (almost) worst-case scenario. $p=.50$ denotes an average-case scenario. Bold font indicates the best result from three independent trials.}
\label{tab:results}
\end{table*}

To provide an explanation for the substantial improvement, we queried each record in the words in context dataset for $25$ times and recorded the cosine similarity between the word pairs after substitution. Since we are only interested in the instances a substitution occurs, we removed cases in which the similarity between substitutions is one. We expect that the similarity between $\hat{w}_i$ and $\hat{w}_j$ obtained from the privatization step is higher when $w_i$ and $w_j$ belong to the same context and lower when different contexts are intended. Whether the words are from an identical context or different contexts is directly derived from annotations. For a transparent comparison, we measure the similarity using \texttt{GloVe} representations of their corresponding substitutions. We present the results in Figure \ref{fig:utility}, separated by word and sense embedding. 

The representations of substitutions obtained by a word embedding convey no clues about the intended contexts the word belongs to. This can be argued by an average similarity that is almost identical at values of $0.1860$ and $0.2035$. Compared to the similarity of lexical representations, the average similarity of substitutions within the same context is $0.3118$ and $0.2272$ for words that originate from different contexts. This distinguishability signals whether words are paired in identical or different contexts, which indicates an awareness of the context during privatization. 

We expect the awareness of the meaning of words to carry over to downstream tasks. To thoroughly evaluate whether context-awareness during privatization translates into better performance on downstream tasks, we conduct experiments on a set of classification tasks in the text domain. We use the General Language Understanding Evaluation (\texttt{GLUE}) benchmark \citep{wang2019glue}. \texttt{GLUE} is a collection of diverse language understanding tasks. The benchmark involves classification of ordinary text and text pairs for similarity and entailment. Apart from \texttt{CoLA} \citep{warstadt2019neural}, which requires high level of syntactic reasoning, all other tasks are based on semantic reasoning. 

We summarize the results on a subset of \texttt{GLUE} obtained by fine-tuning a pre-trained \texttt{BERT} \citep{devlin2019bert} in Table \ref{tab:results}. We report the scores once for word embeddings and once for sense embeddings. Using sense embeddings as opposed to word embedding, the average performance increases from $0.5367$ to $0.6065$. This result confirms our expectation that context awareness during privatization translates into better performances on downstream tasks.

\section{Conclusion}
\label{sec:5}

We redesigned the multivariate mechanism of metric differential privacy in the text domain to account for word meaning during privatization. We accomplished this by replacing the word embedding with a sense embedding and incorporating a sense disambiguation step prior to the noise injection. 

Despite the congestion of the embedding space with senses that stem from the same word form, we experimentally demonstrated that our modification follows the privacy formalization of \citet{feyisetan2020privacy}. Once we recalibrated the privacy budget to ensure plausible deniability, we measured the capability of our mechanism to capture the word meaning. By calculating the similarity of pairs of words in a context that triggers the meaning of each word, we observe that the similarity score for substitutions is consistently higher when both words appear in the same context, and lower when both words appear in different contexts. 

With the confirmation that our mechanism captures word meaning, we were interested in whether the benefits of contextual substitutions translates into superior performance in downstream classification tasks. The results on a set of benchmark datasets demonstrated a substantial boost in generalization performance for tasks that rely on semantic reasoning rather than syntactic reasoning. 

\paragraph{Limitations.} Our modification utilizes sense embeddings. Since the senses were not mapped to an external inventory, the senses cannot be interpreted. Apart from the lack of interpretability, sense embeddings are superseded by contextual embeddings derived from transformer models with sense awareness \citep{huang2019glossbert, levine2020sensebert, scarlini2020sensembert}. While sense embeddings and contextual embeddings are not mutually exclusive, it is necessary to alternate between them for the purpose of privatization and optimization.

\section*{Acknowledgment}
We gratefully acknowledge that this research was supported in part by the \textit{German Federal Ministry of Education and Research} through the \textit{Software Campus} (ref. \textit{01IS17045}).

\bibliography{submission}
\bibliographystyle{acl_natbib}

\end{document}